\pdfoutput=1

\documentclass[11pt]{article}

\usepackage[final]{acl}

\usepackage{times}
\usepackage{latexsym}

\usepackage[T1]{fontenc}

\usepackage[utf8]{inputenc}

\usepackage{microtype}

\usepackage{inconsolata}

\usepackage{graphicx}

\usepackage{subfigure}
\usepackage{amsmath}
\usepackage{amssymb} 
\usepackage{booktabs} 
\usepackage{multirow}
\newcommand{\cQ}{{\cal Q}}

\newcommand{\bits}{b}
\newcommand{\round}{\operatorname{rnd}}
\newcommand{\zeros}{z}
\newcommand{\scale}{s}

\usepackage{tabularx}
\usepackage{adjustbox}

\usepackage{makecell}
\title{``Give Me BF16 or Give Me Death''? \\ 
Accuracy-Performance Trade-Offs in LLM Quantization}

\author{
 \textbf{Eldar Kurtić\textsuperscript{1,2}},
 \textbf{Alexandre Marques\textsuperscript{1}},
 \textbf{Shubhra Pandit\textsuperscript{1}},
 \textbf{Mark Kurtz\textsuperscript{1}},
 \textbf{Dan Alistarh\textsuperscript{1,2}}
\\
\\
 \textsuperscript{1}Red Hat AI,
 \textsuperscript{2}Institute of Science and Technology Austria 
\\
 \small{
   \textbf{Correspondence:} \href{mailto:ekurtic@redhat.com}{ekurtic@redhat.com}, \href{mailto:dalistar@redhat.com}{dalistar@redhat.com}
 }
}

\begin{document}
\maketitle
\begin{abstract}
Quantization is a powerful tool for accelerating large language model (LLM) inference, but the accuracy-performance trade-offs across different formats remain unclear. In this paper, we conduct the most comprehensive empirical study to date, evaluating FP8, INT8, and INT4 quantization across academic benchmarks and real-world tasks on the entire Llama-3.1 model family. Through over 500,000 evaluations, our investigation yields several key findings: (1) FP8 (W8A8-FP) is effectively lossless across all model scales, (2) well-tuned INT8 (W8A8-INT) achieves surprisingly low (1-3\%) accuracy degradation, and (3) INT4 weight-only (W4A16-INT) is more competitive than expected, rivaling 8-bit quantization. 
Further, we investigate the \emph{optimal} quantization format for different deployments by analyzing inference performance through the popular vLLM framework. Our analysis provides clear deployment recommendations: W4A16 is the most cost-efficient for synchronous setups, while W8A8 dominates in asynchronous continuous batching. For mixed workloads, the optimal choice depends on the specific use case. 
Our findings offer practical, data-driven guidelines for deploying quantized LLMs at scale—ensuring the best balance between speed, efficiency, and accuracy. \looseness=-1
\end{abstract}

\section{Introduction}
\label{sec:introduction}

The high computational cost of serving LLMs has driven extensive research into inference acceleration techniques, including quantization~\citep{frantar2022gptq, dettmers2022case, lin2024awq}, speculative decoding~\citep{speculative1, speculative2}, and  pruning~\citep{xia2023sheared, minitron2024}. Among these, quantization—reducing the bitwidth of weights, activations, or both—has emerged as the most widely used approach. However, its key challenge  lies in balancing efficiency and accuracy.

Despite progress, systematic benchmarks and practical deployment guidelines remain scarce. This uncertainty has fueled speculation around quantized models, exemplified by the initial skepticism toward the Llama-3.1-405B quantized model release~\citep{dubey2024llama}, which was later found to be near-lossless in LMSYS Arena user evaluations~\citep{chiang2024chatbot}. To address this gap, we pose the following core question:

\begin{center} \emph{\textbf{What are the practical accuracy-performance trade-offs for popular quantization formats?}}
\end{center}

In this study, we focus on widely supported, computationally efficient quantization formats. Specifically, we examine 8-bit weights and activations (W8A8), using integer (INT) precision for NVIDIA Ampere and older GPUs and floating-point (FP) precision for NVIDIA Hopper and Ada Lovelace. Additionally, we consider 4-bit integer weights with 16-bit activations (W4A16-INT), a competitive low-bit alternative. To evaluate accuracy, we implement a broad automated evaluation suite, spanning both academic and real-world benchmarks. Our academic benchmarks include Open LLM Leaderboard V1~\citep{open-llm-leaderboard-v1} and its more challenging V2 version~\citep{open-llm-leaderboard-v2}, while real-world generative tasks are represented by Arena-Hard-Auto-v0.1~\citep{li2024crowdsourced}, HumanEval~\citep{chen2021codex} and HumanEval+~\citep{evalplus}, and the long-context RULER benchmark~\citep{hsieh2024ruler}. Beyond standard evaluations, we further analyze text similarity between outputs from uncompressed and quantized models to assess generative consistency.

Finally, we conduct an extensive inference performance study, benchmarking vLLM~\citep{kwon2023efficient} (version 0.6.4.post1) across three GPU architectures (A6000, A100, H100) in seven deployment scenarios. Our findings provide a comprehensive view of quantization’s trade-offs and offer practical recommendations for real-world LLM deployment. 
Our main findings are as follows: 

\begin{enumerate}
\item \textbf{W8A8-FP quantization is essentially lossless}, preserving the uncompressed model’s accuracy across all benchmarks, often within the evaluation’s margin of error. This result is achieved with a simple yet robust approach: dynamic per-token activation quantization combined with symmetric weight quantization via round-to-nearest assignment.

\item \textbf{W8A8-INT quantization exhibits only a modest accuracy degradation} of 1–3\% per task on average, far lower than the 10\%+ drops reported in prior work~\citep{li2024evaluating, lee2024comprehensive}. This performance is enabled by dynamic activation quantization or SmoothQuant~\citep{xiao2022smoothquant}, paired with GPTQ~\citep{frantar2022gptq} for symmetric weight quantization.

\item \textbf{W4A16-INT quantization maintains consistently low accuracy loss, performing on par with W8A8-INT}. Surprisingly, we show for the first time that a simple variant of GPTQ outperforms the more recent AWQ method~\citep{lin2024awq} on real-world tasks, challenging prior assumptions about low-bit quantization strategies.

\item \textbf{Beyond accuracy, our text similarity analysis reveals that larger quantized models closely adhere to the word choices and sentence structures of their uncompressed counterparts in autoregressive text generation}. In contrast, smaller quantized models introduce moderate variability in structure but still preserve semantic meaning.

\item \textbf{In terms of performance, \mbox{W4A16-INT} is the most efficient choice for synchronous deployments, while W8A8 formats maximize throughput in asynchronous settings.} The optimal quantization scheme depends on model size, hardware, and deployment needs—whether for latency-sensitive applications like code completion or high-throughput multi-turn chat.
\end{enumerate}

Overall, this work provides the first in-depth study of accuracy vs. performance vs. cost trade-offs for quantized LLMs across formats, algorithms, use cases, and hardware types. We aim for these findings to serve as both a practical deployment guide and a strong and competitive foundation for future research on better quantization techniques.

\section{Background and Related Work}
\label{sec:background}

\subsection{A Primer on Quantization}

Early work focused on INT8 activation quantization and INT4/INT8 weight quantization~\citep{dettmers2022llm, yao2022zeroquant, park2022nuqmm}. A common approach is round-to-nearest (RTN) over groups: given a group of $g$ consecutive weights as a vector $\mathbf{x} \in \mathbb{R}^g$, $b$-bit RTN is defined as:
\vspace{-0.5em}
\begin{align}
\label{eq:quantization}
\cQ(\mathbf{x},\bits) &= \round \Bigg(\frac{\mathbf{x} - \min(\mathbf{x})}{\max(\mathbf{x}) - \min(\mathbf{x})} (2^\bits -1) \Bigg)  \nonumber \\
&= \round((\mathbf{x} - \zeros(\mathbf{x}))/\scale(\mathbf{x})),
\end{align}
\vspace{-1em}

where $\round$ rounds to the nearest integer, $\zeros(\mathbf{x}) = \min(\mathbf{x})$ is the zero point, and $\scale(\mathbf{x}) = (\max(\mathbf{x}) - \min(\mathbf{x})) / (2^\bits -1)$ is the scale, computed using min-max normalization. However, RTN struggles  at INT4 precision and suffers from lossy activation quantization even at INT8~\citep{dettmers2022llm}.

\textbf{Weight Quantization.}
To mitigate weight quantization errors, GPTQ~\citep{frantar2022gptq} introduced second-order weight adjustments using calibration data. Subsequent methods, including AWQ~\citep{lin2024awq}, SqueezeLLM~\citep{kim2023squeezellm}, OWQ~\citep{lee2024owq}, and SpQR~\citep{dettmers2023spqr}, incorporated outlier-aware quantization, storing a fraction of weights in higher precision to enable highly accurate 4-bit quantization. More recent high-compression techniques—QuIP~\citep{chee2023quip}, QuIP\#~\cite{tseng2024quipbetterllmquantization}, QTIP~\citep{tseng2024qtip}, AQLM~\citep{egiazarian2024extreme}, and GPTVQ~\citep{van2024gptvq}—target low-bitwidths using advanced representations such as vector quantization. Yet, these formats are inefficient for batch sizes larger than 1, limiting their practicality. 

\textbf{Activation Quantization.}
Quantizing both weights and activations enables low-bit hardware operations. Yet, activations are difficult to quantize due to \emph{outlier features}—elements up to 100× larger than the average~\citep{dettmers2022llm}. Early attempts extracted outlier columns at runtime, but this is inefficient. SmoothQuant~\citep{xiao2022smoothquant} improves upon this by noticing that outliers are stable across the model and can be precomputed using a calibration set. Follow-up work explored W4A4 quantization~\citep{ashkboos2023towards, ashkboos2024quarotoutlierfree4bitinference} and mixed-precision W4A8~\citep{lin2024qserve, zhang2024qqq}, including KV-cache quantization. While promising, these methods still suffer accuracy loss and lack robust support in high-performance inference frameworks. 

\subsection{Related Work}

A significant body of work has explored the accuracy trade-offs under different quantization schemes~\cite{yao2023zeroquant, liu2023emergent, huang2024good, gong2024makes, li2024evaluating, gong2024llmcbenchmarkinglargelanguage}. However, much of this research relies primarily on academic benchmarks, which do not fully reflect real-world deployment scenarios. Additionally, the lack of hyperparameter tuning in some studies leads to misleading conclusions about  accuracy, as we demonstrate in our experiments. We challenge the claim that 8-bit integer activation quantization causes substantial accuracy degradation~\cite{li2024evaluating, lee2024comprehensive}, providing vast evidence to the contrary.

The closest work to ours is by~\citet{lee2024comprehensive}, which, like most prior studies, focuses on \emph{quantization accuracy}, but overlooks key factors. First, while the authors claim to analyze models up to 405B parameters, they omit open-ended benchmarks at this scale and fail to report full-precision baselines even for academic tasks. Without these references, the impact of quantization remains unclear. To address this, we enable efficient multi-node evaluations for the 405B model, conducting a comprehensive accuracy analysis in both academic and real-world settings.
Second,~\citet{lee2024comprehensive} asserts that AWQ outperforms GPTQ in a 4-bit weight-only quantization setup. We correct this claim, and attribute it to suboptimal hyperparameter choices. Our comparative analysis (Table~\ref{tab:gptq-vs-awq} and Appendix~\ref{app:gptq-vs-awq}) shows that while both methods perform similarly on academic benchmarks, GPTQ exhibits notable gains over AWQ in real-world tasks, particularly coding.

Third, we refute the conclusion that W8A8-INT is significantly inferior to W8A8-FP and W4A16-INT. With proper tuning, W8A8-INT achieves competitive accuracy, with only minor losses. For example, while~\citet{lee2024comprehensive} reports a 10-point accuracy drop for W8A8-INT quantized 405B models on the Open LLM Leaderboard V2 compared to FP8, our approach reduces this to just 0.7 points.

\section{Benchmark Design and Setup}

\subsection{Datasets and Benchmarks}
We categorize benchmarks into three groups: academic, real-world, and text similarity analysis.

\textbf{1. Academic benchmarks}, such as Open LLM Leaderboard V1 and V2~\cite{open-llm-leaderboard-v1, open-llm-leaderboard-v2}, provide structured evaluations for question-answering and reasoning tasks. While widely used for benchmarking, they lack alignment with real-world scenarios involving semantics, variability, and context-awareness. Leaderboard V1 includes tasks like GSM for grade school math~\cite{cobbe2021training}, MMLU and ARC-Challenge for world knowledge and reasoning~\cite{hendrycks2020measuring,clark2018think}, Winogrande and HellaSwag for language understanding~\cite{sakaguchi2021winogrande, zellers2019hellaswag}, and TruthfulQA for factual correctness~\cite{lin2021truthfulqa}. Leaderboard V2 extends this with expert knowledge benchmarks such as MMLU-Pro~\cite{wang2024mmluprorobustchallengingmultitask}, GPQA~\cite{rein2023gpqagraduatelevelgoogleproofqa}, and Big Bench Hard~\cite{suzgun2022challengingbigbenchtaskschainofthought}, as well as multi-step reasoning (MuSR~\cite{sprague2024musrtestinglimitschainofthought}), advanced math (MATH Level 5~\cite{hendrycks2021measuringmathematicalproblemsolving}), and instruction following (IFEval~\cite{zhou2023instructionfollowingevaluationlargelanguage}). By evaluating across both leaderboards, we capture a broad spectrum of reasoning and knowledge domains, using both log-likelihood and text-generation evaluations to stress-test quantized models.

\begin{table*}[t!]
    \setlength{\tabcolsep}{4pt}
    \centering
    \caption{Comparison of GPTQ and AWQ 4-bit weight quantization algorithms (W4A16-INT). We observe a small gap between methods on academic benchmarks (left) but a more pronounced difference in favor of GPTQ on real-world (open-ended) benchmarks (right).}
    \label{tab:gptq-vs-awq}
    \resizebox{\textwidth}{!}{%
    {\small
    \begin{tabular}{lccccccc}
        \toprule
        & \multicolumn{3}{c}{Academic Benchmarks} & \multicolumn{4}{c}{Real-World Benchmarks} \\
        \cmidrule(lr){2-4} \cmidrule(lr){5-8}
        \makecell{Model} & Average Score & Leaderboard V1 & Leaderboard V2 & Average Score & Arena-Hard & HumanEval & MBPP  \\
        \midrule
        Llama-3.1-8B-Instruct & 50.84 & 74.06 & 27.62 & 53.7 & 25.8 & 67.3 & 68.1  \\
        GPTQ~\cite{frantar2022gptq}                  & 49.82 & \textbf{73.11} & 26.53 & \textbf{52.3} & \textbf{24.0} & \textbf{67.1} & \textbf{65.8}  \\
        AWQ~\cite{lin2024awq}                   & \textbf{50.05} & 72.69 & \textbf{27.40}  & 49.4 & 22.3 & 63.0 & 62.8  \\
        \midrule
        Llama-3.1-70B-Instruct & 62.93 & 84.20 & 41.66 & 73.1 & 57.0 & 79.7 & 82.5  \\
        GPTQ~\cite{frantar2022gptq}                   & 62.18 & 83.77 & 40.58 & \textbf{73.1} & \textbf{57.0} & \textbf{80.5} & \textbf{81.9}  \\
        AWQ~\cite{lin2024awq}                    & \textbf{62.53} & \textbf{83.96} & \textbf{41.09} & 72.3 & 56.7 & 79.4 & 80.8  \\
        \bottomrule
    \end{tabular}
    }}
    \vspace{-1em}
\end{table*}

\textbf{2. Real-world benchmarks} evaluate models in practical scenarios such as instruction following, chat, long-context, and code generation. Arena-Hard-Auto-v0.1~\citep{li2024crowdsourced,chiang2024chatbot,arenahard2024} automates LMSYS Chatbot Arena~\citep{chiang2024chatbot} evaluations, using an LLM to judge responses to 500 complex prompts, achieving an 89\% agreement with human rankings~\citep{arenahard2024}. This allows rapid and scalable assessment of chat capabilities without human intervention. For code generation, we evaluate models on HumanEval~\citep{chen2021codex} and its extension HumanEval+~\citep{evalplus}, which test the ability to generate correct and functional code. Finally, we conduct long-context evaluations via the rigorous RULER benchmark~\citep{hsieh2024ruler} which consists of retrieval, multi-hop tracing, information aggregation, and question answering evaluations at sequence lengths from 4k to 128k. 

\textbf{3. Our text similarity analysis} benchmark assesses how closely quantized models' outputs align with their full-precision counterparts. While real-world benchmarks reflect practical usage, their open-ended nature introduces variability, making direct accuracy comparisons challenging. To mitigate this, we analyze output similarity under identical prompts using ROUGE~\citep{lin2004rouge}, BERTScore~\citep{zhang2019bertscore}, and Semantic Textual Similarity (STS)~\citep{reimers-2019-sentence-bert}. ROUGE-1 measures unigram overlap, while ROUGE-L captures structural similarity through the longest common subsequence. BERTScore computes token-level contextual similarity using RoBERTa-large embeddings, and STS assesses semantic alignment at the sentence level via Sentence Transformers built on MiniLM~\citep{wang2020minilm}.

\subsection{Models, Formats, and Algorithms}
We evaluate using the highly-popular Llama 3.1 model series~\cite{dubey2024llama}. To  assess quantization trade-offs, we conduct experiments on the instruction-tuned versions of all available sizes (8B, 70B, and 405B). For each, we examine the three main formats  with kernel support in vLLM: W8A8-FP, W8A8-INT, and W4A16-INT.

\textbf{W8A8-FP} quantizes all linear operators in transformer blocks to an 8-bit floating-point format, using round-to-nearest quantization. Weights follow a symmetric per-output-channel scheme, while activations are dynamically quantized per token. This  requires no calibration data and remains computationally efficient, even for large-scale models.

\textbf{W8A8-INT} reduces weights and activations to 8-bit integers, applying symmetric per-output-channel GPTQ quantization for weights and dynamic per-token quantization for activations. While this scheme performs well for 8B and 405B models, it causes noticeable accuracy drops at 70B. To mitigate this, we apply SmoothQuant, shifting some activation complexity onto weights, which are easier to quantize. For calibration, random tokens suffice at 8B, but larger models require higher-quality calibration data, for which we use~\citet{platypus2023}.

\textbf{W4A16-INT} quantizes weights to 4-bit integers while keeping activations at 16-bit precision. Weights are compressed using GPTQ with MSE-optimal clipping, applied in 128-element groups. Unlike higher-bit formats, random token calibration degrades accuracy, so we rely on OpenPlatypus data for calibration.

\textbf{INT4 Quantization Algorithms.} We focus on two inference-efficient techniques: AWQ and GPTQ, evaluating them on Leaderboard V1/V2, Arena-Hard, HumanEval, and MBPP. Results (Table~\ref{tab:gptq-vs-awq}) show near-identical performance on academic benchmarks, with AWQ leading by just 0.23 and 0.35 points on a 0–100 scale. However, GPTQ outperforms AWQ on real-world tasks by wider margins (2.9 and 0.8 points, respectively), leading us to adopt GPTQ as our primary INT4 method.

This finding contrasts with prior studies~\citep{lin2024awq, huang2024good}, which favored AWQ or found it tied on academic subsets. We attribute this to three key factors: (1) we use GPTQ with MSE-optimal clipping (the AWQ comparison used absmax); this has no overhead and yields consistently better results; (2) we use higher-quality calibration data than the C4 default; (3) we include real-world benchmarks, providing a broader evaluation scope.

\begin{table*}[t!]
    \setlength{\tabcolsep}{4pt}
    \centering
    \caption{Detailed per-task breakdown of accuracy on a subset of academic benchmarks (Open LLM Leaderboard V1) for quantized Llama-3.1-Instruct models across all three model sizes (8B, 70B, 405B). Higher score is better.}
    \label{tab:detailed_leaderboard_v1_for_main_body}
    \resizebox{\textwidth}{!}{%
    {\small 
    \begin{tabular}{llccccccccc}
        \toprule
        & & \makecell{Recovery\\\%} & \makecell{Average \\Score} & \makecell{MMLU \\ 5-shot} & \makecell{MMLU CoT \\ 0-shot} & \makecell{ARC-C \\ 0-shot} & \makecell{GSM8k CoT \\ 8-shot} & \makecell{HellaSwag \\ 10-shot} & \makecell{Winogrande \\ 5-shot} & \makecell{TruthfulQA \\ 0-shot} \\
        \midrule
        \multirow{4}{*}{8B} & BF16 & 100.00 & 74.06 & 68.3 & 72.8 & 81.4 & 82.8 & 80.5 & 78.1 & 54.5 \\
        & W8A8-FP          & \phantom{1}99.31  & 73.55 & 68.0 & 71.6 & 81.2 & 82.0 & 80.0 & 77.7 & 54.3 \\
        & W8A8-INT         & 100.31 & 74.29 & 67.8 & 72.2 & 81.7 & 84.8 & 80.3 & 78.5 & 54.7 \\
        & W4A16-INT        & \phantom{1}98.72  & 73.11 & 66.9 & 71.1 & 80.2 & 82.9 & 79.9 & 78.0 & 52.8 \\
        \midrule
        \multirow{4}{*}{70B} & BF16 & 100.00 & 84.40 & 83.8 & 86.0 & 93.3 & 94.9 & 86.8 & 85.3 & 60.7 \\
        & W8A8-FP           & \phantom{1}99.72  & 84.16 & 83.8 & 85.5 & 93.5 & 94.5 & 86.6 & 84.6 & 60.6 \\
        & W8A8-INT          & \phantom{1}99.87  & 84.29 & 83.7 & 85.8 & 93.1 & 94.2 & 86.7 & 85.1 & 61.4 \\
        & W4A16-INT         & \phantom{1}99.53  & 84.00 & 83.6 & 85.6 & 92.8 & 94.4 & 86.3 & 85.5 & 59.8 \\
        \midrule
        \multirow{4}{*}{405B} & BF16 & 100.00 & 86.79 & 87.4 & 88.1 & 95.0 & 96.0 & 88.5 & 87.2 & 65.3 \\
        & W8A8-FP            & 100.12 & 86.89 & 87.5 & 88.1 & 95.0 & 95.8 & 88.5 & 88.0 & 65.3 \\
        & W8A8-INT           & \phantom{1}99.32  & 86.20 & 87.1 & 87.7 & 94.4 & 95.5 & 88.2 & 86.1 & 64.4 \\
        & W4A16-INT          & \phantom{1}99.98  & 86.78 & 87.2 & 87.7 & 95.3 & 96.3 & 88.3 & 87.4 & 65.3 \\
        \bottomrule
    \end{tabular}
    }}
\end{table*}

\begin{table*}[t!]
    \setlength{\tabcolsep}{2pt}
    \centering
    \caption{Detailed per-task breakdown of accuracy on a subset of academic (Open LLM Leaderboard V2) and on real-world (Arena-Hard, HumanEval, RULER) benchmarks for quantized Llama-3.1-Instruct models across all three model sizes (8B, 70B, 405B). Higher score is better. Long-context RULER evaluations at 405B are prohibitively expensive for our cluster.}
    \label{tab:detailed_leaderboard_v2_for_main_body}
    \resizebox{\textwidth}{!}{%
    {\small
    \begin{tabular}{llcccccccccccc}
        \toprule
        & & \multicolumn{8}{c}{Academic Benchmarks (Open LLM Leaderboard V2)} & \multicolumn{4}{c}{Real-World Benchmarks}\\
        \cmidrule(lr){3-10} \cmidrule(lr){11-14}
        & & \makecell{Recovery\\\%} & \makecell{Average\\Score} & \makecell{IFEval\\0-shot} & \makecell{BBH\\3-shot} & \makecell{Math lvl 5\\4-shot} & \makecell{GPQA\\0-shot} & \makecell{MuSR\\0-shot} & \makecell{MMLU-Pro\\5-shot} & \makecell{Arena-Hard\\Win-Rate} & \makecell{HumanEval\\pass@1} & \makecell{HumanEval+\\pass@1} &
        \makecell{RULER\\Score} \\
        \midrule
        \multirow{4}{*}{8B} & BF16 & 100.0 & 27.6 & 77.8 & 30.1 & 15.7 & 3.7 & 7.6 & 30.8 & 25.8 & 67.3 & 60.7 & 82.8 \\
        & W8A8-FP & 101.2 & 27.9 & 77.2 & 29.6 & 16.5 & 5.7 & 7.5 & 31.2 & 26.8 & 67.3 & 61.3 & 82.8\\
        & W8A8-INT & 101.5 & 28.0 & 77.9 & 30.9 & 15.5 & 5.4 & 7.6 & 30.9 & 27.2 & 67.1 & 60.0 & 82.8\\
        & W4A16-INT & \phantom{1}96.1 & 26.5 & 76.3 & 28.9 & 14.8 & 4.1 & 6.3 & 28.8 & 24.0 & 67.1 & 59.1 & 81.1\\
        \midrule
        \multirow{4}{*}{70B} & BF16 & 100.0 & 41.7 & 86.4 & 55.8 & 26.1 & 15.4 & 18.1 & 48.1 & 57.0 & 79.7 & 74.8 & 83.3\\
        & W8A8-FP & 100.0 & 41.7 & 87.6 & 54.9 & 28.0 & 14.6 & 17.2 & 47.7 & 57.7 & 80.0 & 75.0 & 83.0 \\
        & W8A8-INT & \phantom{1}97.3 & 40.5 & 86.6 & 55.2 & 23.9 & 13.6 & 16.8 & 47.1 & 57.0 & 78.7 & 74.0 & 82.5\\
        & W4A16-INT & \phantom{1}97.4 & 40.6 & 85.7 & 55.0 & 24.4 & 13.8 & 17.2 & 47.2 & 56.3 & 80.5 & 74.2 & 82.2 \\
        \midrule
        \multirow{4}{*}{405B} & BF16 & 100.0 & 48.7 & 87.7 & 67.0 & 38.9 & 19.5 & 19.5 & 59.7 & 67.4 & 86.8 & 80.1 & - \\
        & W8A8-FP & \phantom{1}99.9 & 48.7 & 86.8 & 67.1 & 38.8 & 18.9 & 20.8 & 59.4 & 66.9 & 87.0 & 81.0 & - \\
        & W8A8-INT & \phantom{1}98.3 & 47.9 & 86.9 & 66.7 & 35.8 & 20.4 & 19.2 & 58.4 & 64.6 & 86.9 & 80.4 & -  \\
        & W4A16-INT & \phantom{1}98.9 & 48.2 & 88.0 & 67.5 & 37.6 & 17.5 & 19.4 & 59.3 & 66.5 & 85.1 & 78.9 & -\\
        \bottomrule
    \end{tabular}
    }}
\end{table*}

\section{Quantization Impact on Accuracy}
\label{sec:experiments}

We begin our discussion of the results by examining the accuracy of quantized models across Leaderboard V1 (Table~\ref{tab:detailed_leaderboard_v1_for_main_body}), Leaderboard V2 (Table~\ref{tab:detailed_leaderboard_v2_for_main_body}) and real-world benchmarks (Table~\ref{tab:detailed_leaderboard_v2_for_main_body}). Given the density of the results, we discuss them individually via average recoveries across higher-level benchmarks and discuss ``outlier'' observations. 

\subsection{Academic Benchmarks}
\label{sec:academic}

Our first analysis focuses on Open LLM Leaderboard V1 and V2, ensuring generalization by optimizing quantization hyperparameters on V1 while validating results on V2.

\textbf{The Open LLM Leaderboard V1} follows Meta’s prompt guidelines for Llama-3.1 models to maintain alignment with baseline scores. This introduces two key differences from standard evaluation protocols: MMLU and ARC-Challenge are assessed as text-generation tasks rather than log-likelihood-based evaluations~\cite{eval-harness}, and GSM8k is tested using chain-of-thought prompting instead of a few-shot approach.

\textbf{Table~\ref{tab:detailed_leaderboard_v1_for_main_body} shows that all quantization schemes, across model sizes, recover approximately 99\% of the unquantized BF16 baseline.} The lowest task recovery occurs on TruthfulQA, reaching 96.88\% for W4A16-INT at 8B and  $\sim$98.5\% for larger models (see Appendix Table~\ref{tab:leaderboard_v1_per_task_recoveries}). On average, 8-bit quantization achieves 99.75\%  recovery, while W4A16-INT reaches a   competitive 99.36\%.

\textbf{The Open LLM Leaderboard V2} incorporates more challenging tasks to assess advanced reasoning. Unlike V1, V2 normalizes scores by subtracting the random baseline and rescaling to a 0-100 range, ensuring equal weighting across tasks regardless of inherent difficulty.

\textbf{Table~\ref{tab:detailed_leaderboard_v2_for_main_body} shows that quantized models maintain 99\% of the baseline’s average score, with all models recovering at least 96\%}. However, due to the increased difficulty, smaller models exhibit higher variance, particularly on GPQA and MuSR, where full-precision models already approach random guessing thresholds, reducing the reliability of accuracy recovery signals (Appendix Table\ref{tab:leaderboard_v2_per_task_recoveries}).

Focusing on tasks where the full-precision model scores above 40\%, ensuring a meaningful performance baseline, we observe the lowest per-task recovery for 8-bit FP quantization at 98.44\% on BBH (70B) and for 8-bit INT at 97.8\% on MMLU-Pro (405B). Notably, W4A16-INT models demonstrate superior recovery over W8A8-INT, with a minimum accuracy retention of 98\% for the 8B model on IFEval. This suggests that, for INT, quantizing activations is harder than quantizing weights. 

\begin{figure*}[ht!]
    \centering
    \includegraphics[width=\textwidth]{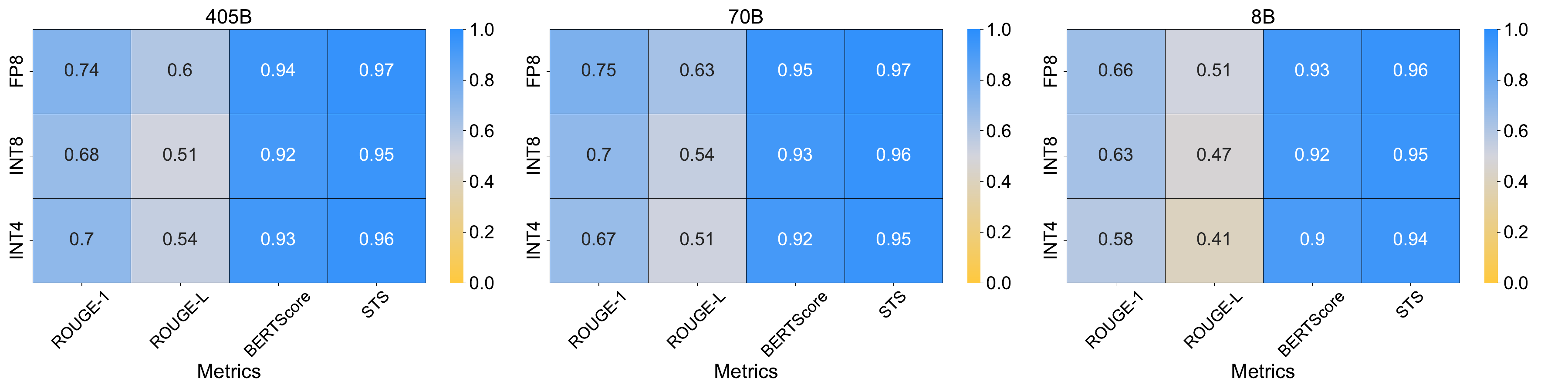}
    \vspace{-2em}
    \caption{Text similarity metrics comparing the outputs of quantized Llama-3.1-Instruct models to full-precision baselines. We refer to W8A8-FP as FP8, W8A8-INT as INT8, and W4A16-INT as INT4.}
    \label{fig:text_similarity}
\end{figure*}

\subsection{Real-World Benchmarks}
\label{sec:realworld}

While academic benchmarks offer structured evaluations, real-world benchmarks better capture model performance in dynamic environments. These evaluations involve diverse prompts, longer generations, and multiple valid responses, emphasizing correctness and semantic quality. We assess four key benchmarks: Arena-Hard-Auto-v0.1 (measuring chat and instruction-following performance, averaging two runs per model and quantization scheme), HumanEval, and HumanEval+ (measuring code generation quality and reporting pass@1 scores using the EvalPlus library~\cite{evalplus}), and RULER (evaluating long-context abilities). Table~\ref{tab:detailed_leaderboard_v2_for_main_body} summarizes the results.

\textbf{On Arena-Hard-Auto-v0.1, quantized models exhibit competitive response quality, with overlapping 95\% confidence intervals across all configurations (Appendix Table~\ref{tab:detailed_arena}).
In coding evaluations, quantized models also maintain strong performance, with 8-bit achieving 99.9\%  recovery and 4-bit recovering 98.9\%, demonstrating their robustness across simple and complex coding tasks}. \textbf{Similarly, for the long-context RULER benchmark, quantized models achieve average score recovery of $\geq$ 98\% across all formats.} 
See Appendix~\ref{app:real_world}  for additional results. 

\subsection{Reasoning Benchmarks}
\label{sec:reasoning}
Given the recent rise in popularity of reasoning abilities of LLMs, we also focus on the popular DeepSeek-R1-Distill~\cite{deepseekai2025deepseekr1incentivizingreasoningcapability} models. These models have been fine-tuned through the process of distillation for improved reasoning capabilities. To assess their reasoning performance, we focus on the challenging and widely recognized reasoning benchmarksi through LightEval~\cite{lighteval}: AIME 2024, MATH-500~\cite{lightman2023let}, and GPQA-Diamond~\cite{rein2024gpqa}. Following DeepSeek’s recommendations for text generation, we use sampling with a temperature of 0.6 and top-p of 0.95, generating 20 responses per query to estimate the pass@1 score. The repetitive sampling was important to estimate an accurate average performance for the benchmarks due to high variance across the relatively small datasets. As can be seen from the results in Table~\ref{tab:reasoning_performance}, the conclusions from the previous sections with academic and real-world benchmarks still hold: \textbf{when quantization is properly tuned and configured, quantized models perform very competitively with their unquantized (BF16) baselines, recovering on average >99\% accuracy except for the smallest models at INT4 which exhibit a bit larger but reasonable drops}.

\begin{table*}[t!]
    \setlength{\tabcolsep}{4pt}
    \centering
    \caption{Detailed per-task and per-model breakdown of accuracy on the popular reasoning benchmarks across all quantized variants of DeepSeek-R1-Distill models from both Llama and Qwen families.}
    \label{tab:reasoning_performance}
    \resizebox{\textwidth}{!}{%
    {\tiny 
    \begin{tabular}{llccccc}
        \toprule
        \multicolumn{2}{c}{DeepSeek-R1-Distill} & \makecell{Recovery\\\%} & \makecell{Average\\Score} & \makecell{AIME24\\pass@1} & \makecell{MATH-500\\pass@1} & \makecell{GPQA-Diamond\\pass@1} \\
        \midrule
        \multirow{4}{*}{Llama-8B} & BF16       & 100.0 & 62.9 & 49.3 ± 6.4 & 90.2 ± 1.2 & 49.3 ± 3.1 \\
        & W8A8-FP    & 100.6 & 63.3 & 50.8 ± 9.0 & 90.2 ± 1.1 & 48.7 ± 2.5 \\
        & W8A8-INT   & 99.6  & 62.7 & 49.1 ± 6.2 & 90.0 ± 1.0 & 48.9 ± 2.0 \\
        & W4A16-INT  & 97.2  & 61.1 & 46.3 ± 6.9 & 89.9 ± 1.1 & 47.1 ± 2.6 \\
        \midrule
        \multirow{4}{*}{Llama-70B} & BF16       & 100.0 & 76.2 & 67.8 ± 7.2 & 95.3 ± 0.7 & 65.6 ± 2.3 \\
        & W8A8-FP    & 100.3 & 76.5 & 69.2 ± 6.5 & 95.1 ± 0.5 & 65.2 ± 2.4 \\
        & W8A8-INT   & 99.7  & 76.0 & 67.8 ± 6.4 & 95.3 ± 0.5 & 65.0 ± 1.8 \\
        & W4A16-INT  & 98.3  & 75.0 & 65.6 ± 5.3 & 95.2 ± 0.6 & 64.0 ± 2.8 \\
        \midrule
        \multirow{4}{*}{Qwen-32B} & BF16       & 100.0 & 76.3 & 69.8 ± 4.9 & 95.1 ± 0.6 & 64.1 ± 2.1 \\
        & W8A8-FP    & 99.0  & 75.6 & 68.5 ± 4.0 & 95.3 ± 0.7 & 62.9 ± 2.6 \\
        & W8A8-INT   & 99.6  & 76.0 & 68.2 ± 5.1 & 95.0 ± 0.8 & 64.8 ± 2.6 \\
        & W4A16-INT  & 99.5  & 75.9 & 68.8 ± 4.2 & 95.0 ± 0.5 & 63.8 ± 1.7 \\
        \midrule
        \multirow{4}{*}{Qwen-14B} & BF16       & 100.0 & 73.6 & 66.7 ± 5.1 & 94.7 ± 0.7 & 59.4 ± 2.3 \\
        & W8A8-FP    & 101.0 & 74.3 & 68.1 ± 5.8 & 94.6 ± 0.6 & 60.1 ± 2.9 \\
        & W8A8-INT   & 99.4  & 73.1 & 66.3 ± 7.1 & 94.7 ± 0.7 & 58.3 ± 2.0 \\
        & W4A16-INT  & 99.0  & 72.8 & 66.0 ± 6.3 & 95.0 ± 0.5 & 57.5 ± 2.1 \\
        \midrule
        \multirow{4}{*}{Qwen-7B} & BF16       & 100.0 & 65.8 & 53.2 ± 6.4 & 93.7 ± 0.8 & 50.5 ± 2.8 \\
        & W8A8-FP    & 99.9  & 65.7 & 53.2 ± 7.5 & 93.6 ± 0.7 & 50.3 ± 2.0 \\
        & W8A8-INT   & 100.7 & 66.3 & 55.2 ± 4.9 & 93.0 ± 1.1 & 50.7 ± 3.5 \\
        & W4A16-INT  & 98.3  & 64.7 & 50.9 ± 7.8 & 93.3 ± 1.1 & 49.8 ± 2.8 \\
        \midrule
        \multirow{4}{*}{Qwen-1.5B} & BF16       & 100.0 & 50.0 & 30.1 ± 5.3 & 84.7 ± 1.1 & 35.4 ± 3.0 \\
        & W8A8-FP    & 100.3 & 50.2 & 29.8 ± 5.6 & 84.7 ± 1.3 & 35.9 ± 3.3 \\
        & W8A8-INT   & 96.9  & 48.5 & 26.7 ± 6.3 & 84.4 ± 1.1 & 34.4 ± 2.8 \\
        & W4A16-INT  & 93.5  & 46.8 & 24.6 ± 5.1 & 82.5 ± 1.1 & 33.2 ± 3.4 \\
        \bottomrule
    \end{tabular}
    }}
\end{table*}

\subsection{Text Similarity Investigation}
\label{sec:textsimilarity}

Next, we analyze the similarity of generated text between quantized and full-precision models. Using Arena-Hard-Auto-v0.1 prompts and greedy sampling for full reproducibility, we compute ROUGE-1, ROUGE-L, BERTScore, and Semantic Textual Similarity (STS) normalized to a 0-1 range.

{As shown in Figure~\ref{fig:text_similarity}, large quantized models (70B and 405B) closely match their full-precision counterparts, achieving an average ROUGE-1 of 0.7 and ROUGE-L of 0.56, indicating strong word and structural preservation}. BERTScore (0.93) and STS (0.96) further confirm semantic consistency despite minor token variations. While 8B models exhibit slightly higher variability, with ROUGE-1 and ROUGE-L dropping to 0.62 and 0.46, they still maintain strong semantic fidelity, as reflected in their BERTScore (0.92) and STS (0.95). \textbf{These results demonstrate that quantized models generate high-quality outputs across all sizes and schemes.}

\begin{table*}[h!]
    \centering
    \caption{Synchronous inference performance comparison across model sizes and GPU configurations. Results show latency (in seconds) and cost-efficiency (Queries per USD) for various tasks. We refer to W8A8-FP as FP8, W8A8-INT as INT8, and W4A16-INT as INT4.}
    \label{tab:synchronous}
    \setlength{\tabcolsep}{3pt}
    \resizebox{\textwidth}{!}{%
    \small  
    \begin{tabular}{l c c c c *{7}{r r}}
        \toprule
        \multirow{2}{*}{\textbf{Size}} & 
        \multirow{2}{*}{\textbf{GPU}} & 
        \multirow{2}{*}{\textbf{\#}} & 
        \multirow{2}{*}{\textbf{Format}} & 
        \multirow{2}{*}{\textbf{CR}} &
        \multicolumn{2}{c}{\textbf{\makecell{Code\\Completion}}} &
        \multicolumn{2}{c}{\textbf{\makecell{Docstring\\Generation}}} &
        \multicolumn{2}{c}{\textbf{\makecell{Code\\Fixing}}} &
        \multicolumn{2}{c}{\textbf{RAG}} &
        \multicolumn{2}{c}{\textbf{\makecell{Instruction\\Following}}} &
        \multicolumn{2}{c}{\textbf{\makecell{Multi-Turn\\Chat}}} &
        \multicolumn{2}{c}{\textbf{\makecell{Summarization}}} \\
        \cmidrule(lr){6-7} \cmidrule(lr){8-9} \cmidrule(lr){10-11} \cmidrule(lr){12-13} 
        \cmidrule(lr){14-15} \cmidrule(lr){16-17} \cmidrule(lr){18-19}
        & & & & & Lat. & Q/\$ & Lat. & Q/\$ & Lat. & Q/\$ & Lat. & Q/\$ & Lat. & Q/\$ & Lat. & Q/\$ & \quad Lat. & {Q/\$\quad} \\
        \midrule
        \multirow{3}{*}{8B} &
        \multirow{3}{*}{A6000} & 1 & BF16 & -- & 24.5 & 183 & 3.2 & 1,395 & 25.0 & 180 & 3.3 & 1,374 & 3.1 & 1,445 & 6.2 & 723 & 13.4 & {335\quad} \\
        & & 1 & INT8 & 1.54	& 15.9 & 284 & 2.1 & 2,157 & 16.3 & 276 & 2.1 & 2,139 & 2.0 & 2,249 & 4.0 & 1,120 & 8.9 & {506\quad} \\
        & & 1 & INT4 & \textbf{2.39} & 9.7 & \textbf{462} & 1.4 & \textbf{3,290} & 10.1 & \textbf{445} & 1.4 & \textbf{3,136} & 1.3 & \textbf{3,543} & 2.5 & \textbf{1,787} & 6.1 & \textbf{736\quad} \\
        \midrule
        \multirow{9}[3]{*}{70B} &
        \multirow{3}{*}{A6000} & 4 & BF16 & -- & 61.7 & 18 & 6.6 & 170 & 62.6 & 18 & 8.1 & 138 & 8.0 & 141 & 15.8 & 71 & 32.6 & {35\quad} \\
        & & 2 & INT8 & 1.94 & 63.4 & 35 & 7.1 & 317 & 63.8 & 35 & 8.4 & 267 & 8.0 & 280 & 16.2 & 139 & 34.0 & {66\quad} \\
        & & 2 & INT4 & \textbf{2.96} & 39.2 & \textbf{57} & 5.0 & \textbf{453} & 40.4 & \textbf{56} & 5.8 & \textbf{390} & 5.1 & \textbf{440} & 10.2 & \textbf{221} & 23.5 & \textbf{96\quad} \\
        \cmidrule(lr){2-19}
        & \multirow{3}{*}{A100} & 2 & BF16 & -- & 50.7 & 20 & 2.9 & 343 & 51.2 & 20 & 6.8 & 148 & 6.4 & 156 & 12.9 & 78 & 27.3 & {37\quad} \\
        & & 1 & INT8 & 1.81 & 54.3 & 37 & 4.0 & 500 & 54.8 & 37 & 7.2 & 279 & 6.9 & 291 & 13.8 & 146 & 29.3 & {69\quad} \\
        & & 1 & INT4 & \textbf{2.67} & 35.0 & \textbf{57} & 2.8 & \textbf{718} & 35.8 & \textbf{56} & 5.2 & \textbf{390} & 4.6 & \textbf{439} & 9.2 & \textbf{220} & 21.0 & \textbf{96\quad} \\
        \cmidrule(lr){2-19}
        & \multirow{3}{*}{H100} & 2 & BF16 & -- & 31.3 & 18 & 4.0 & 139 & 31.5 & 18 & 4.1 & 138 & 4.0 & 142 & 7.9 & 71 & 16.4 & {34\quad} \\
        & & 1 & FP8 & 1.84 & 32.8 & 33 & 4.3 & 256 & 33.1 & 33 & 4.3 & 254 & 4.2 & 262 & 8.3 & 132 & 17.4 & {63\quad} \\
        & & 1 & INT4 & \textbf{2.11} & 28.6 & \textbf{38} & 3.8 & \textbf{289} & 28.2 & \textbf{39} & 3.8 & \textbf{287} & 3.7 & \textbf{299} & 7.1 & \textbf{153} & 15.3 & \textbf{72\quad} \\
        \midrule
        \multirow{6}[2]{*}{405B} &
        \multirow{3}{*}{A100} & 16 & BF16 & -- & 81.9 & 2 & 10.8 & 12 & 81.2 & 2 & 11.2 & 11 & 10.6 & 12 & 20.9 & 6 & 44.1 & {3\quad} \\
        & & 8 & INT8 & 3.27 & 50.1 & 5 & 6.6 & 38 & 50.5 & 5 & 6.8 & 37 & 6.4 & 39 & 12.8 & 20 & 26.9 & {9\quad} \\
        & & 4 & INT4 & \textbf{6.38} & 48.9 & \textbf{10} & 7.0 & \textbf{71} & 49.5 & \textbf{10} & 7.3 & \textbf{68} & 6.4 & \textbf{79} & 12.7 & \textbf{39} & 29.4 & \textbf{17\quad} \\
        \cmidrule(lr){2-19}
        & \multirow{3}{*}{H100} & 16 & BF16 & -- & 50.6 & 1 & 6.5 & 12 & 50.3 & 1 & 6.6 & 11 & 6.4 & 12 & 13.0 & 6 & 26.5 & {3\quad} \\
        & & 8 & FP8 & 3.17 & 31.7 & 5 & 4.2 & 36 & 31.9 & 5 & 4.2 & 36 & 4.1 & 37 & 8.0 & 19 & 16.7 & {9\quad} \\
        & & 4 & INT4 & \textbf{5.15} & 37.5 & \textbf{8} & 5.0 & \textbf{58} & 37.8 & \textbf{8} & 5.1 & \textbf{57} & 4.8 & \textbf{60} & 9.2 & \textbf{32} & 20.4 & \textbf{14\quad} \\
        \bottomrule
        \multicolumn{19}{l}{\textsuperscript{†}\textbf{CR}: Cost Reduction factor compared to BF16 baseline. Higher is better.} \\
        \multicolumn{19}{l}{Lat.: Latency in seconds (lower is better). Q/\$: Queries per USD (higher is better).}
    \end{tabular}
    }
\vspace{-1em}
\end{table*}

\begin{table*}[ht!]
    \centering
    \caption{Asynchronous inference performance evaluation across model sizes and hardware configurations. Results show throughput (queries per second) and cost-efficiency (queries per USD) for various use cases. We refer to W8A8-FP as FP8, W8A8-INT as INT8, and W4A16-INT as INT4.}
    \label{tab:asynchronous}
    \setlength{\tabcolsep}{3pt}
    \resizebox{\textwidth}{!}{
    \begin{tabular}{l c c c *{7}{r r}}
        \toprule
        \multirow{2}{*}{\textbf{Size}} & 
        \multirow{2}{*}{\textbf{HW}} & 
        \multirow{2}{*}{\textbf{Format}} & 
        \multirow{2}{*}{\textbf{Speedup}} &
        \multicolumn{2}{c}{\textbf{\makecell{Code\\Compl.}}} &
        \multicolumn{2}{c}{\textbf{\makecell{Doc.\\Gen.}}} &
        \multicolumn{2}{c}{\textbf{\makecell{Code\\Fixing}}} &
        \multicolumn{2}{c}{\textbf{RAG}} &
        \multicolumn{2}{c}{\textbf{\makecell{Inst.\\Following}}} &
        \multicolumn{2}{c}{\textbf{\makecell{Multi-Turn\\Chat}}} &
        \multicolumn{2}{c}{\textbf{\makecell{Summarization}}} \\
        \cmidrule(lr){5-6} \cmidrule(lr){7-8} \cmidrule(lr){9-10} \cmidrule(lr){11-12} \cmidrule(lr){13-14} \cmidrule(lr){15-16} \cmidrule(lr){17-18}
        & & & & QPS & Q/\$ & QPS & Q/\$ & QPS & Q/\$ & QPS & Q/\$ & QPS & Q/\$ & QPS & Q/\$ & \quad QPS & {Q/\$\quad} \\
        \midrule
        \multirow{3}{*}{8B} &
        \multirow{3}{*}{1×A6000} & BF16 & -- & 1.5 & 6.8k & 5.6 & 25.1k & 1.1 & 4.8k & 4.4 & 19.9k & 11.8 & 53.0k & 5.3 & 24.0k & 0.7 & {3.2k\quad} \\
        & & INT8 & \textbf{1.38} & 2.2 & 9.8k & \textbf{7.7} & \textbf{34.6k} & \textbf{1.4} & \textbf{6.4k} & \textbf{6.1} & \textbf{27.6k} & \textbf{16.5} & \textbf{74.5k} & \textbf{7.2} & \textbf{32.3k} & \textbf{1.0} & \textbf{4.4k\quad} \\
        & & INT4 & 1.08 & \textbf{2.2} & \textbf{9.8k} & 5.3 & 24.0k & 1.3 & 6.0k & 4.1 & 18.6k & 11.2 & 50.5k & 5.4 & 24.3k & 0.7 & {3.1k\quad} \\
        \midrule
        \multirow{9}[3]{*}{70B} &
        \multirow{3}{*}{4×A6000} & BF16 & -- & 0.4 & 0.4k & 1.4 & 1.6k & 0.3 & 0.3k & 1.4 & 1.6k & 3.3 & 3.8k & 1.5 & 1.7k & 0.2 & {0.3k\quad} \\
        & & INT8 & 1.91 & 0.7 & 0.8k & \textbf{3.9} & \textbf{4.4k} & 0.5 & 0.6k & \textbf{2.8} & \textbf{3.1k} & \textbf{6.9} & \textbf{7.7k} & 2.2 & 2.5k & \textbf{0.3} & \textbf{0.4k\quad} \\
        & & INT4 & \textbf{1.92} & \textbf{1.2} & \textbf{1.4k} & 2.7 & 3.1k & \textbf{0.7} & \textbf{0.8k} & 1.9 & 2.1k & 5.2 & 5.9k & \textbf{2.6} & \textbf{3.0k} & 0.3 & {0.3k\quad} \\
        \cmidrule(lr){2-18}
        & \multirow{3}{*}{4×A100} & BF16 & -- & 1.4 & 0.7k & 6.9 & 3.5k & 1.0 & 0.5k & 3.3 & 1.6k & 8.7 & 4.4k & 4.3 & 2.2k & 0.7 & {0.4k\quad} \\
        & & INT8 & \textbf{1.87} & \textbf{2.4} & \textbf{1.2k} & 15.9 & 8.0k & \textbf{1.8} & \textbf{0.9k} & \textbf{6.1} & \textbf{3.1k} & \textbf{16.5} & \textbf{8.3k} & \textbf{8.0} & \textbf{4.0k} & \textbf{1.2} & \textbf{0.6k\quad} \\
        & & INT4 & 1.64 & 2.3 & 1.2k & \textbf{22.8} & \textbf{11.5k} & 1.4 & 0.7k & 4.3 & 2.2k & 11.9 & 6.0k & 5.8 & 2.9k & 0.8 & {0.4k\quad} \\
        \cmidrule(lr){2-18}
        & \multirow{3}{*}{4×H100} & BF16 & -- & 3.5 & 1.0k & 10.0 & 2.9k & 2.6 & 0.7k & 8.0 & 2.3k & 20.3 & 5.9k & 9.9 & 2.9k & 1.7 & {0.5k\quad} \\
        & & FP8 & \textbf{1.77} & \textbf{6.9} & \textbf{2.0k} & \textbf{17.8} & \textbf{5.2k} & \textbf{4.0} & \textbf{1.2k} & \textbf{14.3} & \textbf{4.2k} & \textbf{38.3} & \textbf{11.1k} & \textbf{18.4} & \textbf{5.4k} & \textbf{2.6} & \textbf{0.8k\quad} \\
        & & INT4 & 1.55 & 5.9 & 1.7k & 16.4 & 4.8k & 3.1 & 0.9k & 13.0 & 3.8k & 35.8 & 10.4k & 16.1 & 4.7k & 2.2 & {0.6k\quad} \\
        \midrule
        \multirow{6}[2]{*}{405B} &
        \multirow{3}{*}{16×A100} & BF16 & -- & 0.8 & 59 & 2.5 & 187 & 0.3 & 20 & 2.1 & 156 & 4.6 & 347 & 2.1 & 158 & 0.3 & {22\quad} \\
        & & INT8 & \textbf{2.53} & 1.3 & 98 & \textbf{4.8} & \textbf{358} & 1.1 & 79 & \textbf{3.8} & \textbf{282} & \textbf{10.1} & \textbf{760} & \textbf{4.9} & \textbf{366} & \textbf{0.8} &	\textbf{63\quad} \\
        & & INT4 & 2.21 & \textbf{1.9} & \textbf{144} & 3.6 & 271 & \textbf{1.2} & \textbf{93} & 2.8 & 211 & 8.2 & 616 & 4.0 & 304 & 0.6 & {43\quad} \\
        \cmidrule(lr){2-18}
        & \multirow{3}{*}{16×H100} & BF16 & -- & 0.7 & 52 & 6.1 & 456 & 0.6 & 44 & 4.8 & 363 & 8.5 & 638 & 5.3 & 398 & 0.6 & {46\quad} \\
        & & FP8 & 3.04 & \textbf{4.4} & \textbf{329} & 9.6 & 725 & \textbf{2.7} & \textbf{200} & 7.6 & 571 & 20.7 & 1561 & 10.4 & 780 & \textbf{1.7} & \textbf{125\quad} \\
        & & INT4 & \textbf{3.09} & 4.0 & 304 & \textbf{11.1} & \textbf{833} & 2.5 & 192 & \textbf{8.7} & \textbf{652} & \textbf{24.7} & \textbf{1856} & \textbf{11.6} & \textbf{872} & 1.6 & {122\quad} \\
        \bottomrule
        \multicolumn{18}{l}{QPS: Queries per second (higher is better). Q/\$: Queries per USD (higher is better).} \\
        \multicolumn{18}{l}{Numbers denoted with \textit{k} represent thousands (e.g., 20.3k = 20,300).}
    \end{tabular}
    }
\vspace{-1em}
\end{table*}

\section{Quantized Inference Performance}
\label{sec:inference_performance}
LLM inference consists of two main stages: prefill, where all input tokens are processed simultaneously, and decode, where tokens are generated sequentially. Prefill is typically compute-bound, while decode is memory-bound.
Weight quantization primarily accelerates decode by reducing memory movement, whereas weight-and-activation quantization improves computational efficiency in prefill. Thus, the optimal choice for quantization scheme depends on the ratio of prefill to decode tokens.
Beyond direct speedups, quantization also enhances end-to-end performance by increasing the number of simultaneous queries, improving efficiency, and enabling lower-cost GPU usage for memory-constrained tasks. Thus, real-world deployment involves complex trade-offs.

To assess these trade-offs, we benchmarked W8A8-FP, W8A8-INT, and W4A16-INT across three GPU types (A6000, A100, H100) in seven use cases. Tasks like code completion and instruction following involve short prefill phases (256 tokens) and varying decode lengths (1024 and 128 tokens, respectively). More complex tasks like summarization require significantly longer prefill (4096 tokens) with a moderate decode length (512 tokens). Multi-turn chat and RAG involve moderate prefill lengths (512 and 1024 tokens) with shorter decode phases (256 and 128 tokens). Finally, docstring generation (768 prefill, 128 decode) and code fixing (1024 prefill, 1024 decode) reflect intermediate token requirements. For latency-sensitive applications, we compare both synchronous and asynchronous deployment under latency constraints, while throughput-driven cases are evaluated in asynchronous mode. To assess cost efficiency across hardware setups, we use Lambda Labs' on-demand GPU pricing~\cite{lambda_labs}, shown in Table~\ref{tab:lambda_labs}, which is standard.

\subsection{Synchronous Deployment}

Latency-sensitive applications are sometimes deployed in synchronous mode, where a single query is processed at a time. This approach minimizes latency by avoiding resource contention, making inference largely decode-bound.

Table~\ref{tab:synchronous} compares inference performance across model sizes, GPU types, quantization schemes, and use cases, highlighting the most cost-effective GPU configurations. The results show that W4A16-INT consistently achieves the highest performance gains across all models and hardware setups.

For 8B and 70B models, W4A16-INT reduces cost per query by 2–3× and improves latency by 1.5–2.5× compared to the full-precision BF16 baseline. The impact is even more pronounced at 405B, where W4A16-INT achieves 5–7× cost reductions and enables inference with fewer GPUs. Notably, deploying the 405B model on 4× A100 or H100 GPUs with W4A16-INT meets performance thresholds that previously required 16 GPUs in BF16, reducing inter-GPU communication and latency. Given the minor accuracy trade-offs observed in the previous section, this makes W4A16-INT highly effective for synchronous deployment.

\subsection{Asynchronous Deployment}

Processing multiple queries concurrently improves computational efficiency compared to single-query execution. vLLM automatically manages asynchronous requests, balancing computation between prefill and decode stages.

While asynchronous deployment increases per-query latency relative to synchronous execution, it amortizes computation across multiple requests, significantly boosting overall throughput, measured in queries per second (QPS). Table~\ref{tab:asynchronous} reports the maximum achievable throughput and cost efficiency (queries per dollar) across different quantization formats, model sizes, and hardware configurations. The setups were optimized for peak BF16 performance and kept consistent when evaluating quantized models. \textbf{Results show that W8A8-INT and W8A8-FP yield the highest throughput, though W4A16-INT remains competitive and can outperform W8A8 in some scenarios.}

\begin{figure}[h!]
    \centering
    \includegraphics[width=\columnwidth]{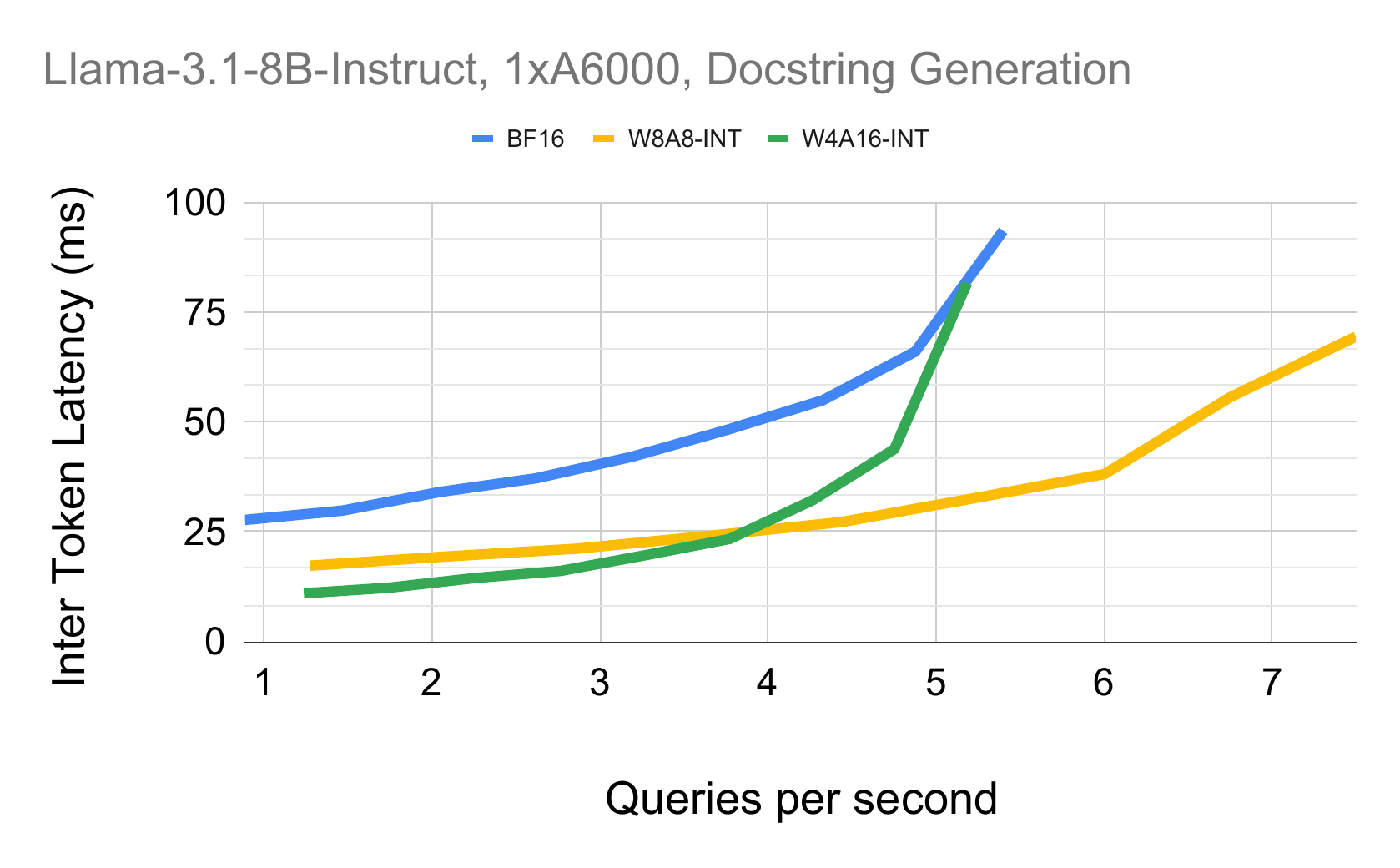}
    \caption{Latency-throughput example for docstring generation use-case. W4A16 is more efficient at low latency (lower throughput), whereas W8A8 becomes more efficient at high latency (high throughput).}
    \label{fig:async1}
\end{figure}

\begin{figure}[h!]
    \centering
    \includegraphics[width=\columnwidth]{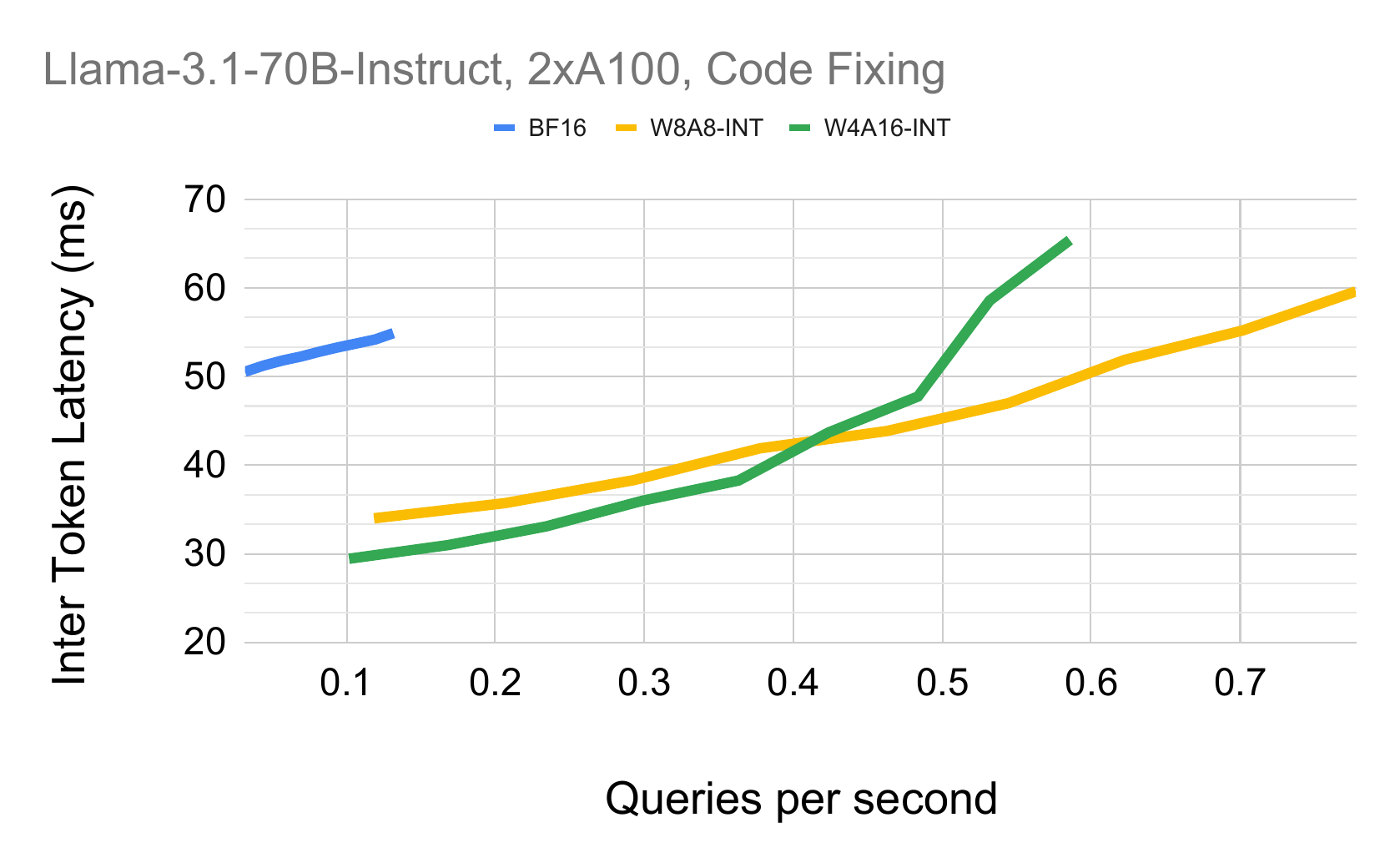}
    \caption{Latency-throughput example for code fixing use-case. W4A16 is more efficient at low latency (lower throughput), whereas W8A8 becomes more efficient at high latency (high throughput).}
    \label{fig:async2}
\end{figure}

Many real-world applications impose latency constraints on asynchronous deployment. Figures~\ref{fig:async1} and ~\ref{fig:async2} illustrate trade-offs between latency and throughput for two example tasks: docstring generation and code fixing. \textbf{W4A16-INT is more efficient at lower latencies, making it ideal for applications requiring rapid response times. In contrast, W8A8 formats maximize throughput at the cost of higher latency, making them better suited for batch processing}. The point where W8A8 overtakes W4A16  depends on factors such as model size, hardware, and task requirements.

\begin{figure}[h!]
    \centering
    \includegraphics[width=0.95\columnwidth]{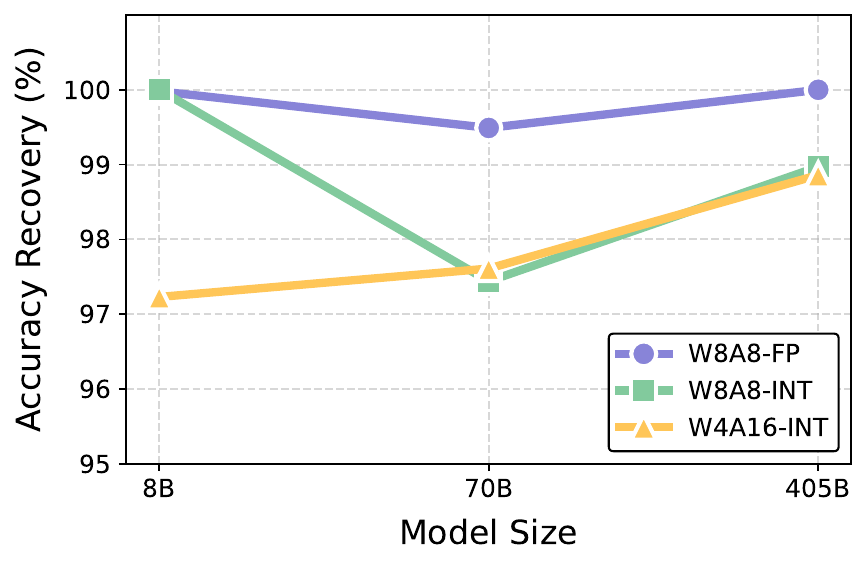}
    \vspace{-1em}
    \caption{Accuracy recovery trends across academic benchmarks highlight the challenges of integer activation quantization, particularly at larger model sizes.}
    \label{fig:quant_trends}
\end{figure}

\section{Conclusion}
\label{sec:conclusion}
We provided a broad, in-depth study of \mbox{accuracy-vs-performance-vs-cost} trade-offs for quantized LLMs across various deployment environments, covering all quantization formats with efficient support, and a range of quantization algorithms, deployment use cases, and GPUs. In Figure~\ref{fig:quant_trends} we summarize our findings in terms of accuracy recovery per quantization format, using  carefully-tuned state-of-the-art quantization techniques.
Broadly, our findings show that, with a judicious choice of algorithm and parametrization, these formats can offer higher accuracy than previously thought, significantly improve inference performance, and reduce costs. At the same time, we have also shown that the optimal choice of format can be task and algorithm specific, providing guidelines for this choice. 

\section*{Limitations}

While our study provides a comprehensive evaluation of quantization effects on model accuracy and inference performance, several limitations remain. We have primarily focused on weight and activation quantization, leaving open questions about the impact of compressing other model components such as the KV-cache, input embeddings, and language modeling head. Further investigation is needed to assess how these additional compression techniques influence both accuracy and computational efficiency. Additionally, our analysis does not fully explore the effects of quantization across specialized use cases, such as multi-lingual tasks, where accuracy degradation could vary significantly depending on the language distribution and underlying model architecture. Future work should extend these evaluations to provide a more holistic understanding of quantization trade-offs in diverse deployment scenarios.

\section*{Acknowledgements}
This research was funded in whole or in part by the Austrian Science Fund (FWF) 10.55776/COE12.

\bibliography{custom}

\appendix

\section{Additional Results}
\subsection{Real-World Benchmarks}
\label{app:real_world}

\begin{figure}[h!]
    \centering
    \includegraphics[width=\columnwidth]{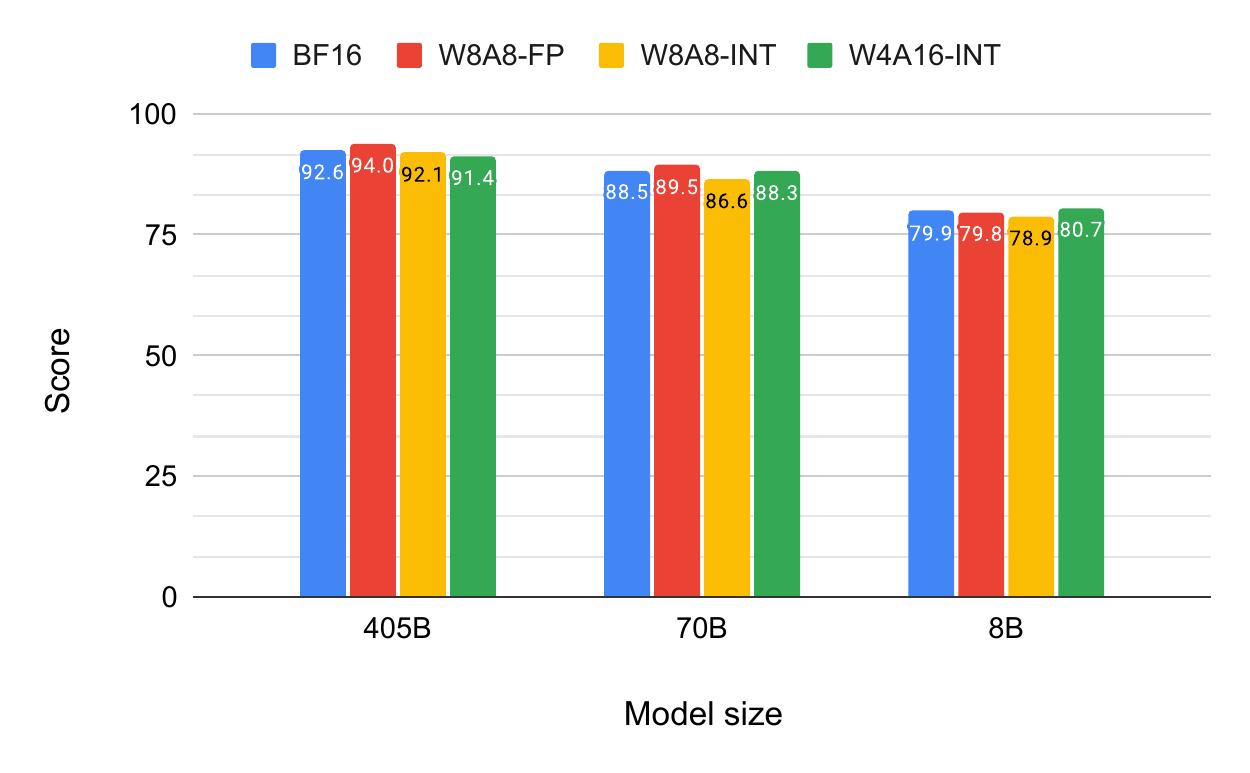}
    \caption{HumanEval pass@10 scores for quantized Llama-3.1-Instruct models.}
    \label{fig:humaneval_pass10}
\end{figure}

\begin{figure}[h!]
    \centering
    \includegraphics[width=\columnwidth]{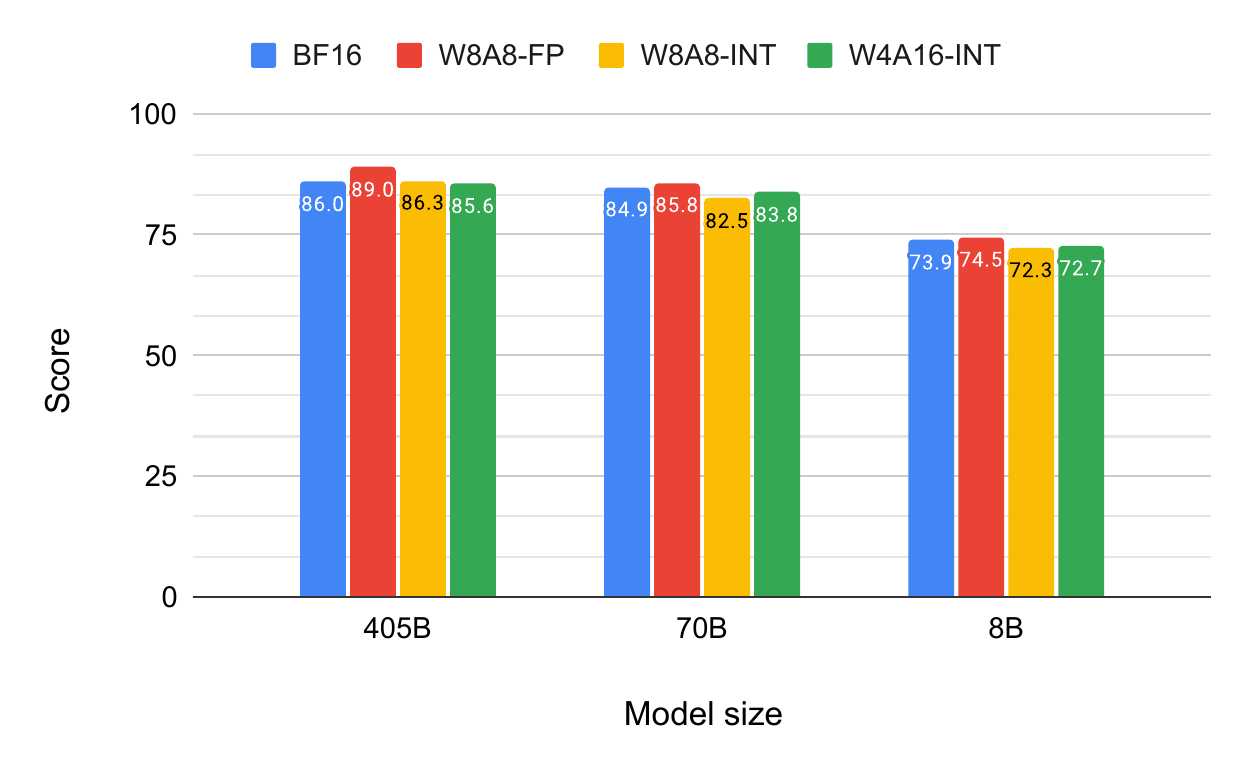}
    \caption{HumanEval+ pass@10 scores for quantized Llama-3.1-Instruct models.}
    \label{fig:humanevalplus_pass10}
\end{figure}

In Figures~\ref{fig:humaneval_pass10} and~\ref{fig:humanevalplus_pass10} we report pass@10 scores for all models on HumanEval and HumanEval+ benchmarks.

In Table~\ref{tab:detailed_arena} we report scores of two Arena-Hard-Auto-v0.1 runs, aggregated average scores, and 95\% confidence intervals (CI).

\begin{table}[ht!]
    \centering
    \caption{Scores and confidence intervals of two evaluation runs for Llama-3.1-Instruct models through Arena-Hard-Auto-v0.1.}
    \label{tab:detailed_arena}
    \small
    \begin{tabular}{lcccc}
        \toprule
        \makecell{Llama-3.1\\Instruct} & \makecell{Score\\(1st run)} & \makecell{Score\\(2nd run)} & \makecell{Average\\Score} & 95\% CI \\
        \midrule
        BF16 405B & 67.3 & 67.5 & 67.4 & (-2.6, 1.9) \\
        W8A8-FP & 66.3 & 67.55 & 66.9 & (-2.6, 2.3) \\
        W8A8-INT & 64.3 & 64.8 & 64.6 & (-2.4, 2.8) \\
        W4A16-INT & 66.5 & 66.4 & 66.5 & (-2.6, 2.3) \\
        \midrule
        BF16 70B & 55.8 & 58.2 & 57.0 & (-2.6, 2.1) \\
        W8A8-FP & 57.6 & 57.75 & 57.7 & (-2.4, 3.1) \\
        W4A16-INT & 57.1 & 56.8 & 57.0 & (-2.8, 2.5) \\
        W8A8-INT & 56.0 & 56.6 & 56.3 & (-2.9, 2.4) \\
        \midrule
        BF16 8B & 25.1 & 26.5 & 25.8 & (-2.1, 2.1) \\
        W8A8-FP & 26.8 & 26.85 & 26.8 & (-2.1, 2.6) \\
        W8A8-INT & 27.6 & 26.7 & 27.2 & (-2.0, 2.2) \\
        W4A16-INT & 23.4 & 24.6 & 24.0 & (-2.2, 2.0) \\
        \bottomrule
    \end{tabular}
\end{table}

\subsection{Detailed Comparison of GPTQ and AWQ}
\label{app:gptq-vs-awq}
To complement the results in Table~\ref{tab:gptq-vs-awq}, Tables~\ref{tab:gptq-vs-awq-arena-hard},~\ref{tab:gptq-vs-awq-leaderboard-v1},~\ref{tab:gptq-vs-awq-leaderboard-v2} provide a detailed per-task and per-run breakdown of scores.

\begin{table}[h!]
    \setlength{\tabcolsep}{4pt}
    \centering
    \caption{Comparison of GPTQ and AWQ quantization algorithms, both with group size of 128, across two runs of the Arena-Hard-Auto-v0.1 benchmark.}
    \label{tab:gptq-vs-awq-arena-hard}
    \small
    \begin{tabular}{lccc}
        \toprule
        & \makecell{Score \\ (1st run)} & \makecell{Score \\ (2nd run)} & \makecell{Average\\Score} \\
        \midrule
        Llama-3.1-70B-Instruct & 55.8 & 58.2 & 57.0 \\
        GPTQ~\cite{frantar2022gptq}                   & 57.1 & 56.8 & 57.0 \\
        AWQ~\cite{lin2024awq} & 56.3 & 57.0 & 56.3 \\
        \midrule
        Llama-3.1-8B-Instruct  & 25.1 & 26.5 & 25.8 \\
        GPTQ~\cite{frantar2022gptq} & 23.4 & 24.6 & 24.0 \\
        AWQ~\cite{lin2024awq}                    & 22.4 & 22.2 & 22.3 \\
        \bottomrule
    \end{tabular}
\end{table}

\subsection{GPU Pricing}
We use Lambda Labs' on-demand GPU pricing~\cite{lambda_labs}, as displayed in Table~\ref{tab:lambda_labs}.
For A100 GPUs Lambda Labs only provides the 8x configuration. For scenarions with a smaller number of A100 GPUs we assume a price proportional to the number of GPUs. 

\begin{table}[h!]
    \setlength{\tabcolsep}{4pt}
    \centering
    \caption{On-demand hardware cost on Lambda Labs' cloud.}
    \label{tab:lambda_labs}
    \begin{tabular}{lc}
        \toprule
        Hardware & \makecell{On-demand cost\\(USD per hours)}\\
        \midrule
        1xA6000 & 0.80\\
        2xA6000 & 1.60\\
        4xA6000 & 3.20\\
        \midrule
        8xA100 & 14.32\\
        \midrule
        1xH100 & 3.29\\
        2xH100 & 6.38\\
        4xH100 & 12.36\\
        8xH100 & 23.92\\
        \bottomrule
    \end{tabular}
\end{table}

\subsection{Academic Benchmarks}
\label{app:academic}
In Tables~\ref{tab:leaderboard_v1_per_task_recoveries} and~\ref{tab:leaderboard_v2_per_task_recoveries} we report accuracy recoveries per-task across academic benchmarks.

\begin{table*}[h!]
    \centering
    \caption{Accuracy recoveries in percentages (\%) for each task in the Open LLM Leaderboard V1 benchmark.}
    \label{tab:leaderboard_v1_per_task_recoveries}
    {\small
    \begin{tabular}{lccccccc}
        \toprule
        \makecell{Llama-3.1-Instruct} & \makecell{MMLU \\ 5-shot} & \makecell{MMLU CoT \\ 0-shot} & \makecell{ARC-C \\ 0-shot} & \makecell{GSM8k CoT \\ 8-shot} & \makecell{HellaSwag \\ 10-shot} & \makecell{Winogrande \\ 5-shot} & \makecell{TruthfulQA \\ 0-shot} \\
        \midrule
        Baseline BF16 8B & 100.00 & 100.00 & 100.00 & 100.00 & 100.00 & 100.00 & 100.00 \\
        W8A8-FP          & 99.59  & 98.35  & 99.75  & 99.03  & 99.38  & 99.49  & 99.63 \\
        W8A8-INT         & 99.27  & 99.18  & 100.37 & 102.42 & 99.75  & 100.51 & 100.37 \\
        W4A16-INT        & 97.95  & 97.66  & 98.53  & 100.12 & 99.25  & 99.87  & 96.88 \\
        \midrule
        Baseline BF16 70B & 100.00 & 100.00 & 100.00 & 100.00 & 100.00 & 100.00 & 100.00 \\
        W8A8-FP           & 100.00 & 99.42  & 100.21 & 99.58  & 99.77  & 99.18  & 99.84 \\
        W8A8-INT          & 99.88  & 99.77  & 99.79  & 99.26  & 99.88  & 99.77  & 101.15 \\
        W4A16-INT         & 99.76  & 99.53  & 99.46  & 99.47  & 99.42  & 100.23 & 98.52 \\
        \midrule
        Baseline BF16 405B & 100.00 & 100.00 & 100.00 & 100.00 & 100.00 & 100.00 & 100.00 \\
        W8A8-FP            & 100.11 & 100.00 & 100.00 & 99.79  & 100.00 & 100.92 & 100.00 \\
        W8A8-INT           & 99.66  & 99.55  & 99.37  & 99.48  & 99.66  & 98.74  & 98.62 \\
        W4A16-INT          & 99.77  & 99.55  & 100.32 & 100.31 & 99.77  & 100.23 & 100.00 \\
        \bottomrule
    \end{tabular}
    }
\end{table*}

\begin{table*}[h!]
    \centering
    \caption{Accuracy recoveries in percentages (\%) for each task in the Open LLM Leaderboard V2 benchmark.}
    \label{tab:leaderboard_v2_per_task_recoveries}
    \small
    \begin{tabular}{lcccccc}
        \toprule
        Llama-3.1-Instruct & \makecell{IFEval \\ 0-shot} & \makecell{BBH \\ acc\_norm \\ 3-shot} & \makecell{Math lvl 5 \\ exact\_match \\ 4-shot} & \makecell{GPQA \\ acc\_norm \\ 0-shot} & \makecell{MuSR \\ acc\_norm \\ 0-shot} & \makecell{MMLU-Pro \\ acc \\ 5-shot} \\
        \midrule
        Baseline BF16 8B & 100.00 & 100.00 & 100.00 & 100.00 & 100.00 & 100.00 \\
        W8A8-FP                    & 99.10  & 98.54  & 105.42 & 155.98 & 98.82  & 101.33 \\
        W8A8-INT                   & 100.12 & 102.89 & 98.92  & 146.20 & 100.00 & 100.26 \\
        W4A16-INT                  & 98.00  & 96.08  & 94.39  & 109.78 & 83.18  & 93.63 \\
        \midrule
        Baseline BF16 70B & 100.00 & 100.00 & 100.00 & 100.00 & 100.00 & 100.00 \\
        W8A8-FP                     & 101.34 & 98.44  & 107.52 & 94.68  & 94.49  & 99.13 \\
        W8A8-INT                    & 100.17 & 98.89  & 91.83  & 88.38  & 92.62  & 97.86 \\
        W4A16-INT                   & 99.22  & 98.60  & 93.52  & 89.94  & 94.99  & 98.19 \\
        \midrule
        Baseline BF16 405B & 100.00 & 100.00 & 100.00 & 100.00 & 100.00 & 100.00 \\
        W8A8-FP                      & 99.00  & 100.12 & 99.69  & 97.38  & 106.93 & 99.43 \\
        W8A8-INT                     & 99.20  & 99.57  & 91.94  & 104.51 & 98.77  & 97.81 \\
        W4A16-INT                    & 100.39 & 100.73 & 96.53  & 89.85  & 99.54  & 99.35 \\
        \bottomrule
    \end{tabular}
\end{table*}

\begin{table*}[h!]
    \setlength{\tabcolsep}{4pt}
    \centering
    \caption{Comparison of GPTQ and AWQ quantization algorithms, both with group size of 128, across Open LLM Leaderboard V1 benchmarks~\cite{open-llm-leaderboard-v1} with Meta's prompts~\cite{dubey2024llama}.}
    \label{tab:gptq-vs-awq-leaderboard-v1}
    {\small
    \begin{tabular}{lcccccccc}
        \toprule
        & Average Score & \makecell{MMLU \\ 5-shot} & \makecell{MMLU CoT \\ 0-shot} & \makecell{ARC-C \\ 0-shot} & \makecell{GSM8k CoT \\ 8-shot} & \makecell{HellaSwag \\ 10-shot} & \makecell{Winogrande \\ 5-shot} & \makecell{TruthfulQA \\ mc2 \\ 0-shot} \\
        \midrule
        Llama-3.1-8B-Instruct & 74.06 & 68.30 & 72.80 & 81.40 & 82.80 & 80.50 & 78.10 & 54.50 \\
        GPTQ~\cite{frantar2022gptq}                  & 73.11 & 66.90 & 71.10 & 80.20 & 82.90 & 79.90 & 78.00 & 52.80 \\
        AWQ~\cite{lin2024awq}                   & 72.69 & 66.37 & 69.76 & 80.89 & 82.56 & 79.61 & 76.80 & 52.81 \\
        \midrule
        Llama-3.1-70B-Instruct & 84.20 & 82.37 & 86.06 & 93.30 & 94.90 & 86.80 & 85.30 & 60.70 \\
        GPTQ~\cite{frantar2022gptq}                   & 83.77 & 82.03 & 85.54 & 92.80 & 94.40 & 86.30 & 85.50 & 59.80 \\
        AWQ~\cite{lin2024awq}                    & 83.96 & 82.15 & 85.64 & 93.00 & 94.47 & 86.44 & 85.79 & 60.23 \\
        \bottomrule
    \end{tabular}
    }
\end{table*}

\begin{table*}[h!]
    \setlength{\tabcolsep}{4pt}
    \centering
    \caption{Comparison of GPTQ and AWQ quantization algorithms, both with group size of 128, across Open LLM Leaderboard V2 benchmarks~\cite{open-llm-leaderboard-v2}.}
    \label{tab:gptq-vs-awq-leaderboard-v2}
    \small
    \begin{tabular}{lccccccc}
        \toprule
        & Average Score & \makecell{IFEval \\ 0-shot} & \makecell{BBH \\ acc\_norm \\ 3-shot} & \makecell{Math lvl 5 \\ exact\_match \\ 4-shot} & \makecell{GPQA \\ acc\_norm \\ 0-shot} & \makecell{MuSR \\ acc\_norm \\ 0-shot} & \makecell{MMLU-Pro \\ acc \\ 5-shot} \\
        \midrule
        Llama-3.1-8B-Instruct & 27.62 & 77.86 & 30.09 & 15.68 & 3.68 & 7.61 & 30.77 \\
        GPTQ~\cite{frantar2022gptq}                  & 26.53 & 76.30 & 28.91 & 14.80 & 4.04 & 6.33 & 28.81 \\
        AWQ~\cite{lin2024awq}                   & 27.40 & 78.25 & 27.20 & 13.87 & 5.21 & 10.45 & 29.41 \\
        \midrule
        Llama-3.1-70B-Instruct & 41.66 & 86.41 & 55.79 & 26.07 & 15.40 & 18.16 & 48.12 \\
        GPTQ~\cite{frantar2022gptq} & 40.58 & 85.74 & 55.01 & 24.38 & 13.85 & 17.25 & 47.25 \\
        AWQ~\cite{lin2024awq}                    & 41.09 & 86.60 & 55.24 & 25.14 & 13.68 & 18.81 & 47.06 \\
        \bottomrule
    \end{tabular}
\end{table*}

\end{document}